 \documentclass[smallcondensed]{svjour3}     
\smartqed  
\input{psfig.sty}
\usepackage{epsfig}
\usepackage{amsmath}
\usepackage{amssymb}
\usepackage{bbold}
\usepackage[round]{natbib}	
\begin{document}

\title{An analysis of numerical issues in neural training by pseudoinversion 
}

\titlerunning{An analysis of numerical issues in neural training by pseudoinversion}        

\author{R.~Cancelliere   \and   R.~Deluca \and
        M.~Gai   \and P.~Gallinari \and  L.~Rubini    
}


\institute{R. Cancelliere, R. Deluca, L. Rubini \at
              University of Turin, Department of Computer Sciences, Turin, Italy \\
              Tel.: +39 011 6706737, Fax: +39 011 751603\\
              \email{rossella.cancelliere@unito.it}           
           \and
           M. Gai \at
              National Institute of Astrophysics, Astrophysical Observatory of Torino, Turin, Italy
	\and
	P. Gallinari  \at
	Laboratory of Computer Sciences, LIP6, Universit\'e Pierre et Marie Curie, Paris, France
}

\date{Received: date / Accepted: date}

\maketitle

\begin{abstract}

Some novel strategies have recently been proposed  for single hidden layer neural
network training that set randomly the weights from input to hidden
layer, while weights from hidden to output layer are analytically determined
by pseudoinversion.
These techniques are gaining 
popularity in spite of their known numerical issues when singular and/or almost singular matrices 
are involved. In this paper we discuss a critical use of Singular Value 
Analysis for identification of these drawbacks and we propose an original use of regularisation to determine the output weights, based on the concept of critical hidden layer size. This approach also allows to limit the training computational effort.
Besides, we introduce a novel technique which relies an effective determination of input weights to the hidden layer dimension.
This approach is tested for both regression and classification tasks, resulting in a significant 
performance improvement with respect to alternative methods. 


\end{abstract}

\section{Introduction}
\label{intro}

The training of one of the most common neural architecture, the single hidden
layer feedforward neural network (SLFN) was mainly accomplished in past decades by methods based on gradient descent, and among them the large family of techniques based on backpropagation \citep{rumel}. 
The start-up of these techniques assigns random values to the weights connecting input, hidden
and output nodes that are then iteratively modified according
to the error gradient steepest descent direction. Some common drawbacks with
gradient descent-based learning are anyway the high computational
cost because of slow convergence and the relevant risk of converging to poor local minima on the
landscape of the error function \citep{lecun}.

The idea of using the simple and efficient training algorithms of radial basis function neural networks, based on matricial pseudoinversion \citep{poggio1}, also for SLFN learning was initially suggested in \citep{cancelliere}; some appealing techniques were than developed \citep{nguyen,kohno,ajorloo} and among them the extreme learning machine ELM, \citep{huang} which has been
successfully applied to a number of real-world applications \citep{Sun,wang,malathi,minhas},
showing a good generalization performance with an extremely fast
learning speed.

ELM main result states that SLFNs with randomly chosen input weights and hidden layer biases can learn distinct observations with a desired precision, provided that activation functions in the hidden layer are infinitely differentiable.

After input weights and hidden layer biases have been randomly set, output weights are directly evaluated by Moore-Penrose generalised inverse (or pseudoinverse) of the hidden layer output matrix: these two steps conclude one training phase and weights are no more modified, so that their determination is no more iterative in the sense of back-propagation based techniques.

Besides, all pseudoinversion based methods are multi-start, i.e. the above procedure is repeated meny times in order to find a good minimum of the error surface. Each training procedure so implies many random settings of input weights and as many evaluations of output weights through pseudoinversion. 

However, such techniques seem to require more hidden units than 
typical values from backpropagation training to achieve comparable accuracy, as discussed in Yu and Deng \citep{yu}. 
Moreover, pseudoinversion, commonly evaluated by Singular Value Decomposition 
(SVD), is a powerful method but some caution is required, since its numerical 
instability is a well known issue when singular and almost singular matrices 
are involved. 

One aim of this paper is the analysis of these instability issues; a preliminary assessment of the context 
and our initial results are discussed in \citep{cancelliere2}. 
Here we present further advances on the theoretical framework and we propose a novel approach to carry out a more efficient learning, showing how singular values of SVD can be used to detect the occurrence of numerical instability.

Besides, we prove the existence of a critical hidden layer dimension that allows a careful tuning of the regularisation parameter and the use of regularisation to replace unstable, ill-posed problems with well-posed ones. We also propose an original method to set input weights that links their size to the hidden layer dimension and we show its effectiveness. 

In section \ref{pseudo} we introduce the notation used for describing SLFN architectures, and the main ideas concerning input weights setting and output weights evaluation by pseudoinversion. In section \ref{regul} we discuss the problem of ill-posedness, and the basic regularisation concepts. 

In section \ref{res} our framework is tested on 
some applications selected from the UCI database. 
A substantial improvement in performance with respect to unregularised 
state-of-the-art techniques is shown.

\section{Input and output weights determination}
\label{pseudo}

\begin{figure}[h]
\centering
\psfig{figure=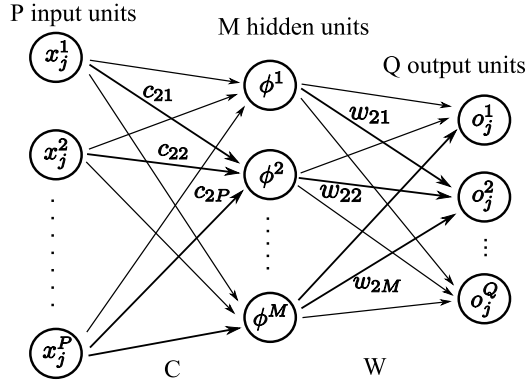,height=5cm}
\caption{A Single Layer Feedforward Neural Network}
\label{fig:SLFN}
\end{figure}

Fig. \ref{fig:SLFN} shows a standard SLFN with $P$ input neurons, $ M$ hidden neurons with non-linear activation functions $ \phi $, and $Q$ output neurons with linear activation functions.

If we have a training set made by $N$ distinct training samples of (input, output) pairs $ (\textbf {x}_j, \textbf {t}_j)$, where $ \textbf {x}_j \in  \mathbb{R}^P $ and $ \textbf {t}_j \in  \mathbb{R}^Q $, the training aims at obtaining  the matrix of desired outputs  $T  \in  \mathbb{R}^{N \times Q} $ when the matrix of all input instances $X  \in  \mathbb{R}^{N \times P}$ is presented as input.

We emphasise that, in the state of the art pseudoinverse approach, input weights $ c_{ij} $ (and hidden neurons biases) are randomly sampled from a uniform distribution in a fixed interval and no 
longer modified. Therefore this step gives the actual input weights values.

After having fixed the input weights matrix C, the use of linear output units allows to determine output weights $w_{ij}$ as the solution of the linear system 

\begin{equation}
H \, W \, = T,
\label{eq:3}
\end{equation}

where  $H \in  \mathbb{R}^{N \times M} $ is the hidden layer output matrix of the neural network, $ H=\Phi (X \, C) $. It is important to underline that because usually the number of hidden nodes is much lower than the number of distinct training samples, i.e. $ M<<N $, $ H$  is a rectangular matrix. 

The least-squares solution  $ \bar{W} $ of the linear system (\ref{eq:3}), as shown e.g. in \citep{penrose,bishop}, is $ \bar{W}=H^+ T, $ where $ H^+ $ is the Moore-Penrose generalised inverse (or pseudoinverse) of matrix $ H $. It can be computed in a computationally simple and accurate way by using singular value decomposition (SVD) \citep{golub}.

We know that every matrix $H \in  \Re^{N \times M} $ can be decomposed as follows:
\begin{equation}
H=U \Sigma V^T \, ,
\label{eq:6}
\end{equation}
where $ U \in  \Re^{N \times N}$  and $ V \in  \Re^{M \times M} $ are orthogonal matrices and $ \Sigma \in  \Re^{N \times M} $ is a rectangular diagonal matrix whose elements $ \sigma _{ii} \equiv \sigma_i $, called singular values, are nonnegative (usually the singular values are listed in descending order, i.e. $ \sigma_1 \geq \sigma_2 \geq \cdots \geq \sigma_p \geq 0 $, $ p$ = min $ \{ N,M \} $, so that $ \Sigma $ is uniquely determined).

The pseudoinverse matrix $ H^+ $ has the form 
\begin{equation}
H^+ =V \Sigma^+ U^T , 
\label{eq:7}
\end{equation}
where $ \Sigma^+ $ is obtained from $ \Sigma $ by taking the reciprocal of each non-zero element $  \sigma _i $, and transposing the resulting matrix \citep{rao}. The presence of very small element $  \sigma _i $ is therefore a potential drawback of this method.

For computational reasons, the elements $ \sigma_i $ equal to zero or smaller than a predefined threshold are replaced by zeros \citep{golub}.

\section{Singular value decomposition of regularised problems}
\label{regul}
 
To turn an original ill-posed problem into a well-posed one, i.e. roughly speaking into a problem insensitive to small changes in training conditions, regularisation methods are often used \citep{badeva}, and Tikhonov regularisation is one of the most common \citep{tikhonov1,tikhonov2}.

The error functional to be minimised is characterized by a penalty term $ E_R $ that depends from the so called Tikhonov matrix $\Gamma $:
\begin{equation}
E\equiv E_D + E_R=|| H W-T||_2^2 + ||\Gamma W ||_2^2,
\label{eq:reg1}
\end{equation}

This matrix can for instance derive from the choice of using  highpass operators 
(e.g. a difference operator or a weighted Fourier operator) 
to enforce smoothness. 

The regularised solution, that we denote by $ \hat{W}$ , is now given by:
\begin{equation}
\hat{W} = (H^T H+ \Gamma^{T} \Gamma )^{-1}H^{T}T \, . 
\label{eq:reg2}
\end{equation}

The penalty term improves 
on stability, making the problem less sensitive to initial conditions, 
and contains model complexity avoiding overfitting, 
as largely discussed in \citep{gallinari}. 

To give preference to solutions $ \hat{W} $ with smaller norm \citep{bishop} a frequent choice is $ \Gamma = \sqrt \lambda I $, so eqs. (\ref{eq:reg1}) and (\ref{eq:reg2}) can be rewritten as
\begin{equation}
E\equiv E_D + E_R=|| H W-T||_2^2 + \lambda || W ||_2^2, 
\label{eq:reg3}
\end{equation}
\begin{equation}
\hat{W} = (H^T H+ \lambda I )^{-1}H^{T}T \, . 
\label{eq:reg4}
\end{equation}

\noindent 
The role of the control parameter $ \lambda$ is to 
trade off between the two error terms $E_D$ and $E_R$. If $ \lambda = 0$, eq.(\ref{eq:reg4}) reduces to the unregularised least-squares solution, provided that $ (H^T H)^{-1} $ exists. 

The regularised solution  (\ref{eq:reg4}) can also be expressed 
(see e.g. \citep{fuhry}) as:
\begin{equation}
\hat{W} = V D U^T T 
\label{eq:tikmat}
\end{equation}
where $ V , U $ are from the singular value decomposition of $ H $ (eq.(\ref{eq:6})) and $ D $ is a rectangular diagonal matrix with elements

\begin{equation}
 D_{i} = \frac{\sigma _i}{\sigma _i ^2 + \lambda} \, . 
\label{eq:tiksing}
\end{equation}
\noindent  obtained using the singular values of $ H $.

It is evident that, when unregularised pseudoinversion is used, the presence of very small singular values can easily causes numerical instability in $ H^+ $; on the contrary, regularisation has a dramatic impact because, even 
in presence of very small values $\sigma_i$ of the original unregularised 
problem, a careful choice of the parameter $ \lambda $ allows to tune 
singular values $D_{i}$ of the regularised matrix, preventing them 
from divergence. 

It is clear at this point that a suitable value for the parameter $ \lambda$ has to derive from a compromise between the necessity to have it sufficiently large to control the approaching to zero of $\sigma$ in eq.(\ref{eq:tiksing}) while avoiding predominance of penalty term in eq.(\ref{eq:reg3}). 
Its tuning is therefore crucial to simultaneously control numerical instability and overfitting.

In the next section we propose a strategy to obtain this result showing that 
we achieve better performance and more stable solutions. 

\section{Experiments and Results}
\label{res}

Some numerical instability issues have already been evidenced in our previous investigations 
\citep{cancelliere2}; we provided suggestions on possible mitigation techniques 
like selection of a convenient activation function and normalisation of the input 
weights. 
Hereafter we show that adding regularisation to the implementation prescriptions already analysed provides a convenient and effective approach to deal with such problem. 

The use of sigmoidal activation functions has recently been subject of debate 
because they seem to be more easily driven towards saturation due to their non-zero mean value \citep{glorot}, while hyperbolic tangent seems less sensitive to this problem.

We select therefore both activation functions for a test aiming at comparing their performance in a context where we also compare our proposed regularised approach and the unregularised one. Four different experimental settings will be analysed, namely HypT-reg, Sigm-reg, HypT-unreg and Sigm-unreg.

\begin{table}
\caption{The UCI datasets and their characteristics }
\label{tab:datasets}       
\begin{tabular}{lllll}
\hline\noalign{\smallskip}
Dataset & Type & N. Instances & N. Attributes & N. Classes \\
\noalign{\smallskip}\hline\noalign{\smallskip}
Abalone & Regression & 4177 & 8 & - \\
Cpu & Regression & 209 & 6 & - \\
Delta Aileron & Regression & 7129 & 5 & - \\
Iris & Classification & 150 & 4 & 3 \\
Diabetes & Classification & 768 & 8 & 2 \\
Landsat & Classification & 4435 & 36 & 7 \\
\noalign{\smallskip}\hline
\end{tabular}
\end{table}

To further mitigate saturation issues, in our previous work \citep{cancelliere2} we selected input weights according to a uniform random distribution in the range ($-1/ \sqrt M$, $1/ \sqrt M$),  where $M$ is the number of hidden nodes.
This links the size of input weights, and therefore of hidden neurons inputs, to the network architecture,  thus forcing the use of the almost linear central part of the hyperbolic and sigmoidal activation functions when exploring the performance as a function of an increasing number of nodes. 
Such prescriptions are retained in the current work. 

We emphasise that so doing input weights are automatically chosen "small", because the size of interval ($-1/ \sqrt M$, $1/ \sqrt M$) decreases when the number of hidden neurons $ M $ increases: for instance with 10 hidden neurons, weights values are roughly selected in the range $ (-1/3, 1/3) $, with 100 hidden neurons in the range $ (-1/10, 1/10) $, and so on.

Because of the wide use among researchers belonging to ELM-community (see for instance \citep{helmy,huang,Sun} 
our performance is also compared with that from unregularised pseudoinversion, input weights selected according to a random uniform distribution in the interval $ (-1, 1) $ and sigmoidal activation functions (hereafter, ELM). 

The numerical experiment compares these frameworks applying them to six benchmark datasets from the UCI repository 
\citep{Bache}, listed in Table \ref{tab:datasets}. 

For each proposed method the number of hidden nodes is gradually increased 
by unity steps, and, for each selected size of SLFN, average RMSE or 
average misclassification rate on the validation set are computed over a set 
of 100 simulation trials, i.e. over 100 different initial 
choices of input weights. 
All simulations are carried out in Matlab 7.3 environment.  

\begin{figure}
\centering
\includegraphics[width=0.48\textwidth]{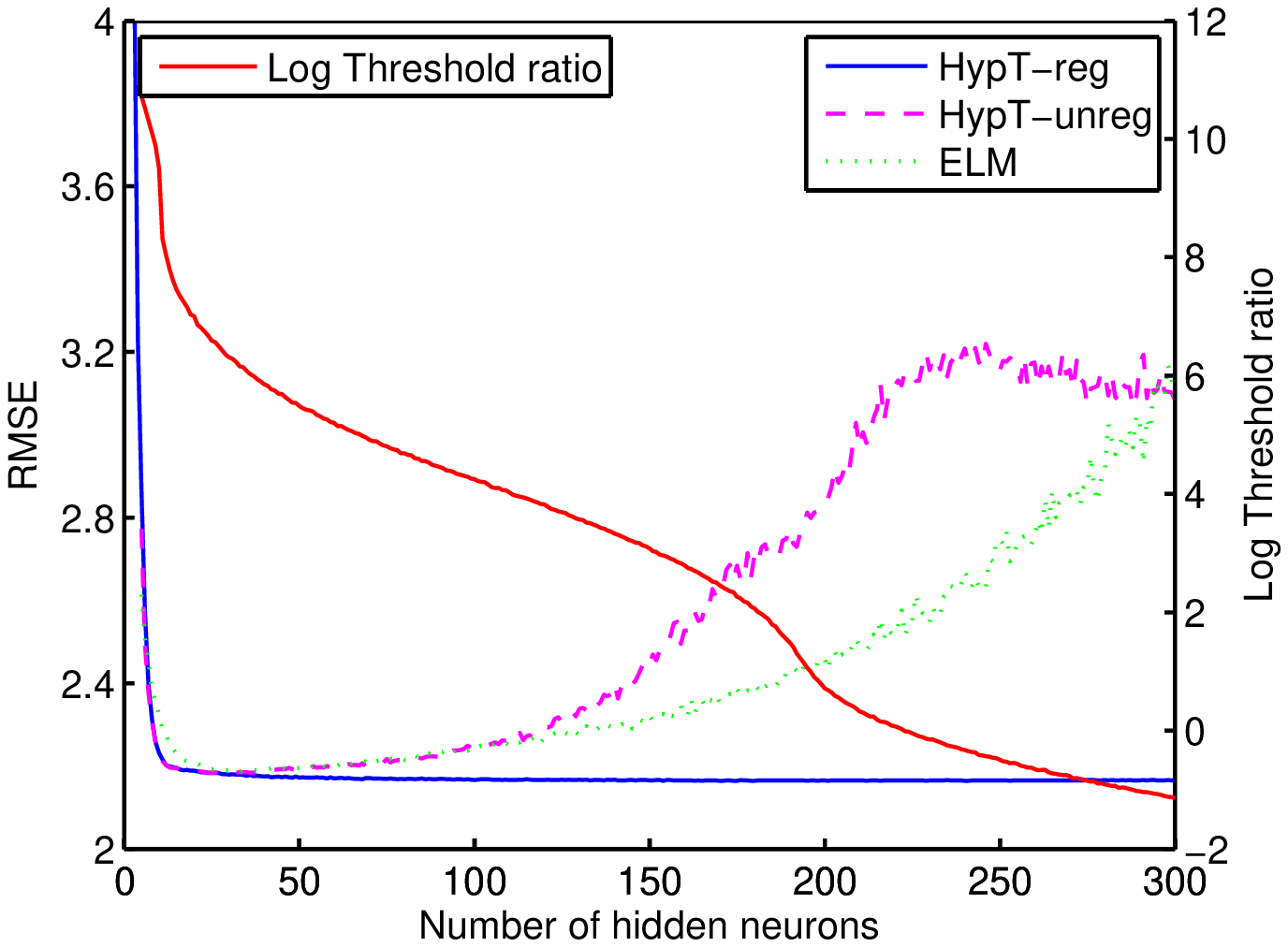}
\includegraphics[width=0.48\textwidth]{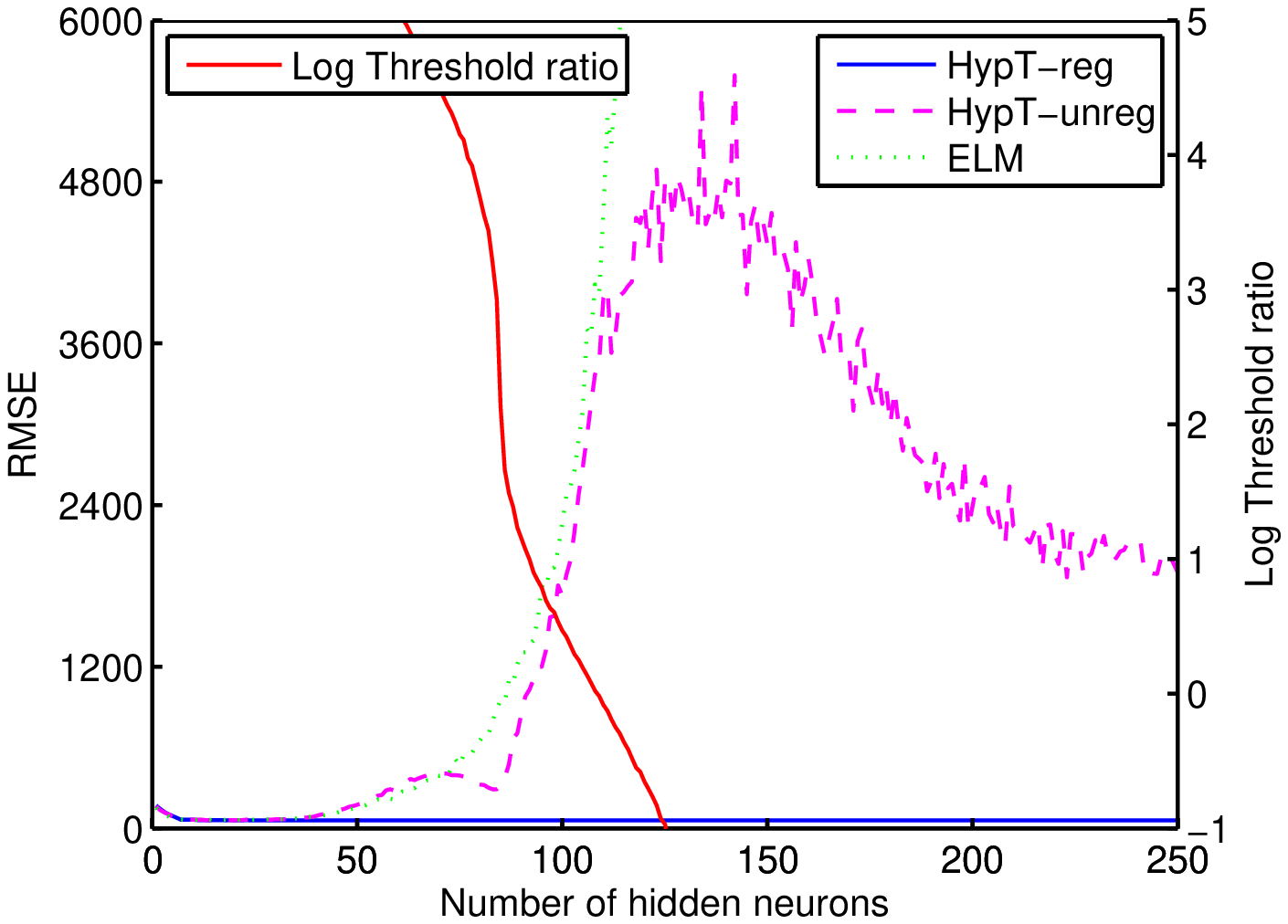}

\includegraphics[width=0.48\textwidth]{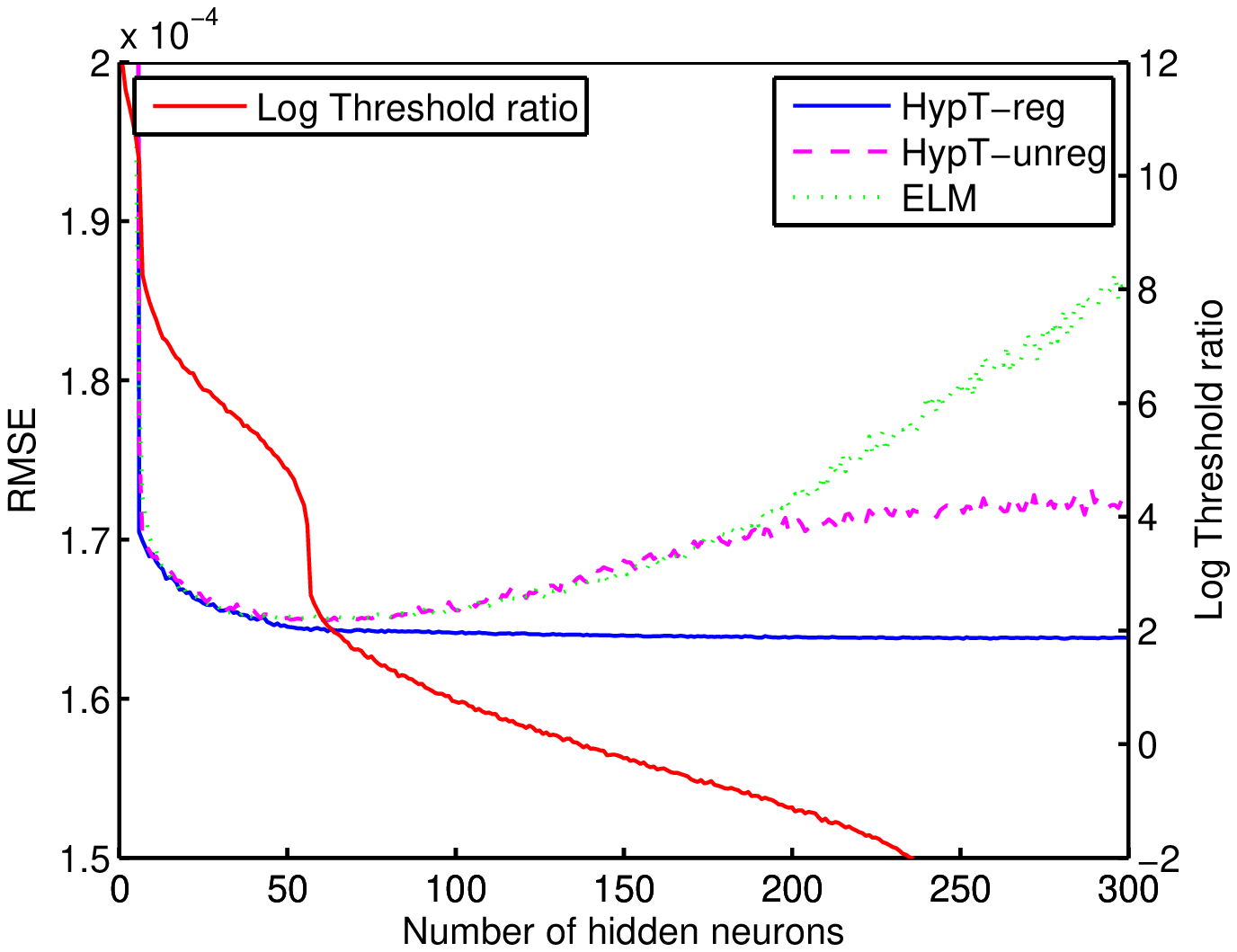}
\includegraphics[width=0.48\textwidth]{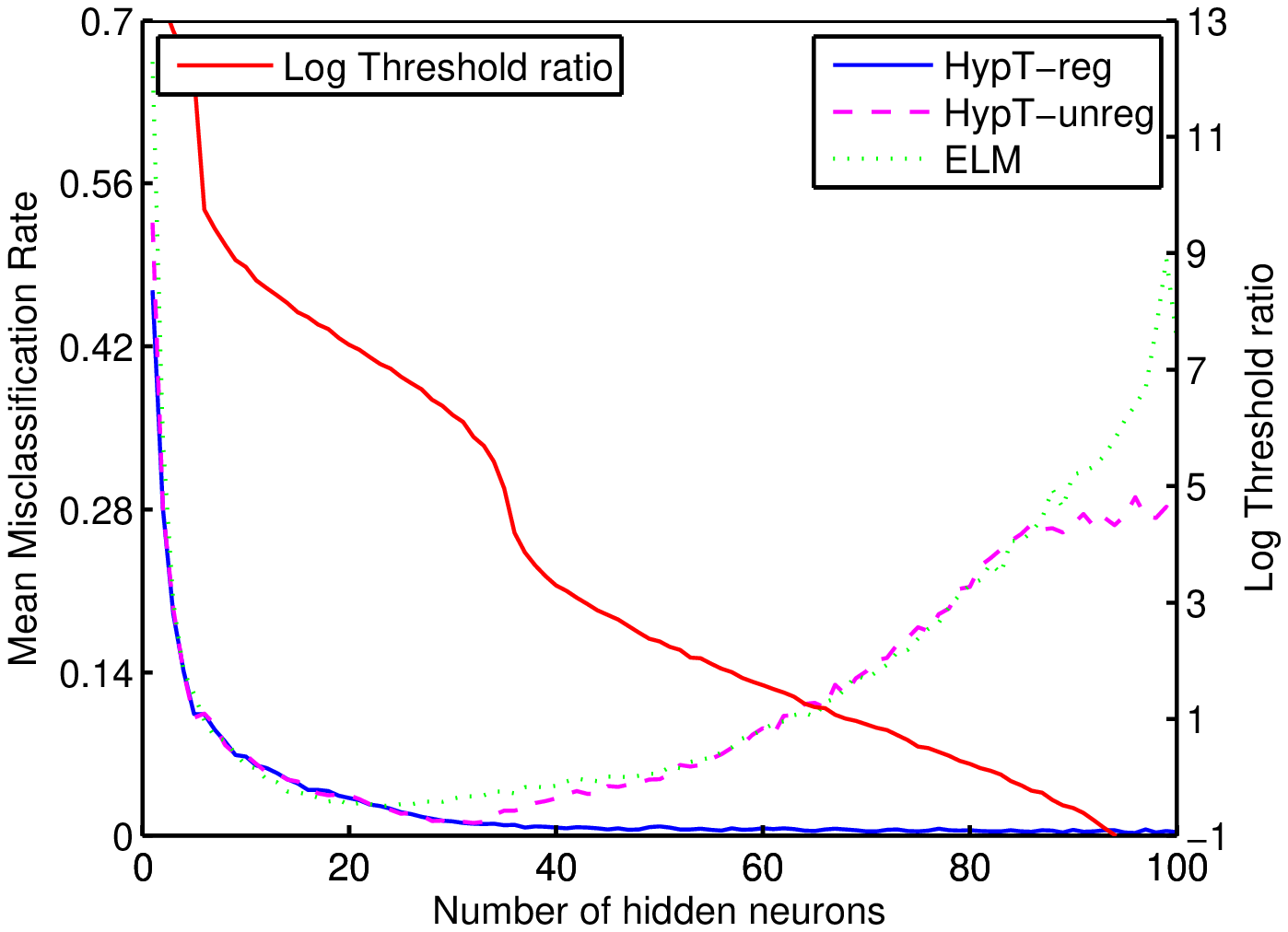}

\includegraphics[width=0.48\textwidth]{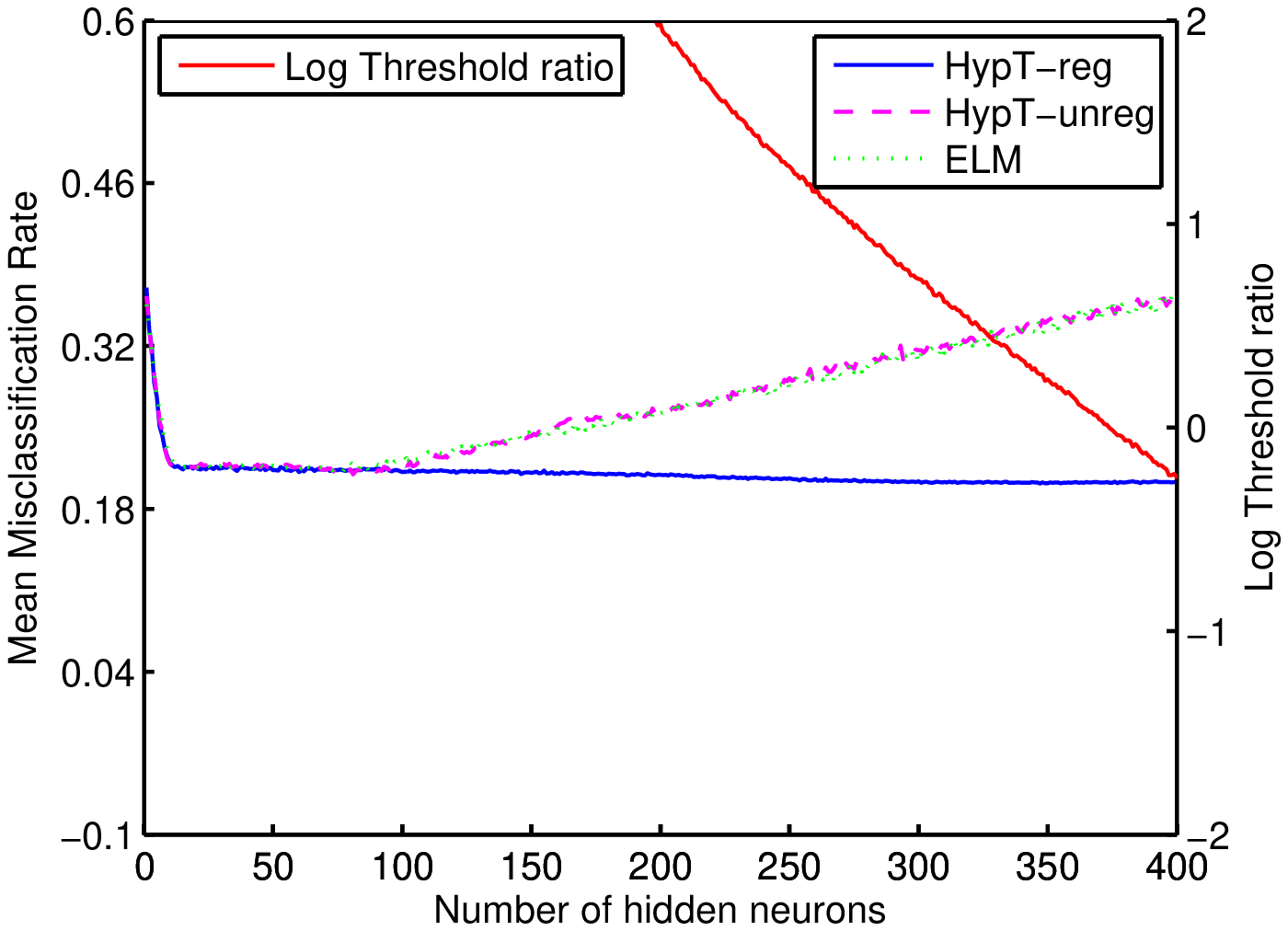}
\includegraphics[width=0.48\textwidth]{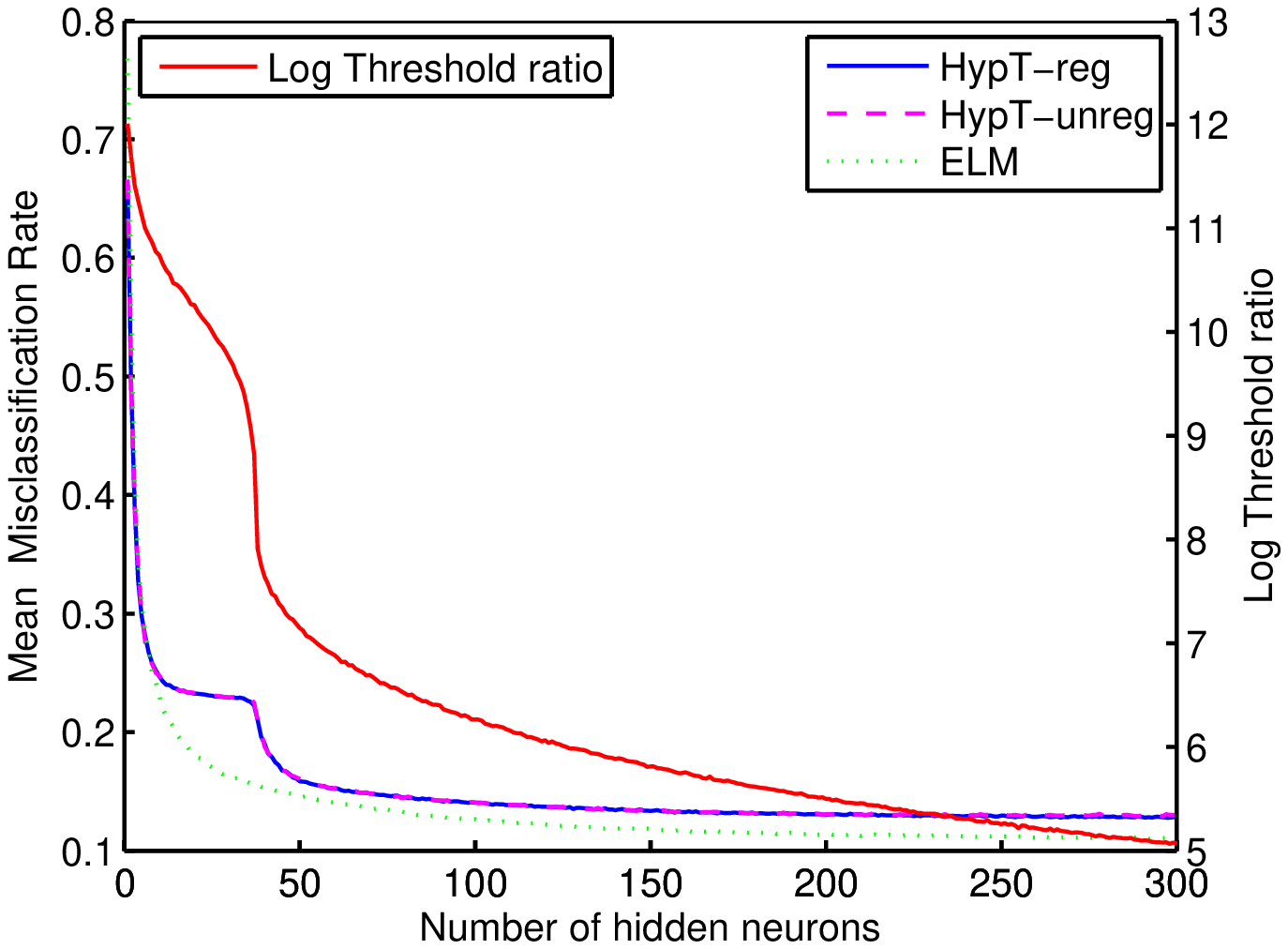}

\caption{Performance comparison in the HypT (-reg and -unreg) cases: Abalone (Top Left);
Cpu (Top Right); Delta Ailerons (Mid Left); Iris (Mid Right);
Diabetes (Bottom Left); Landsat (Bottom Right). }
\label{fig:PlotArray}
\end{figure}

\begin{figure}
\centering
\includegraphics[width=0.48\textwidth]{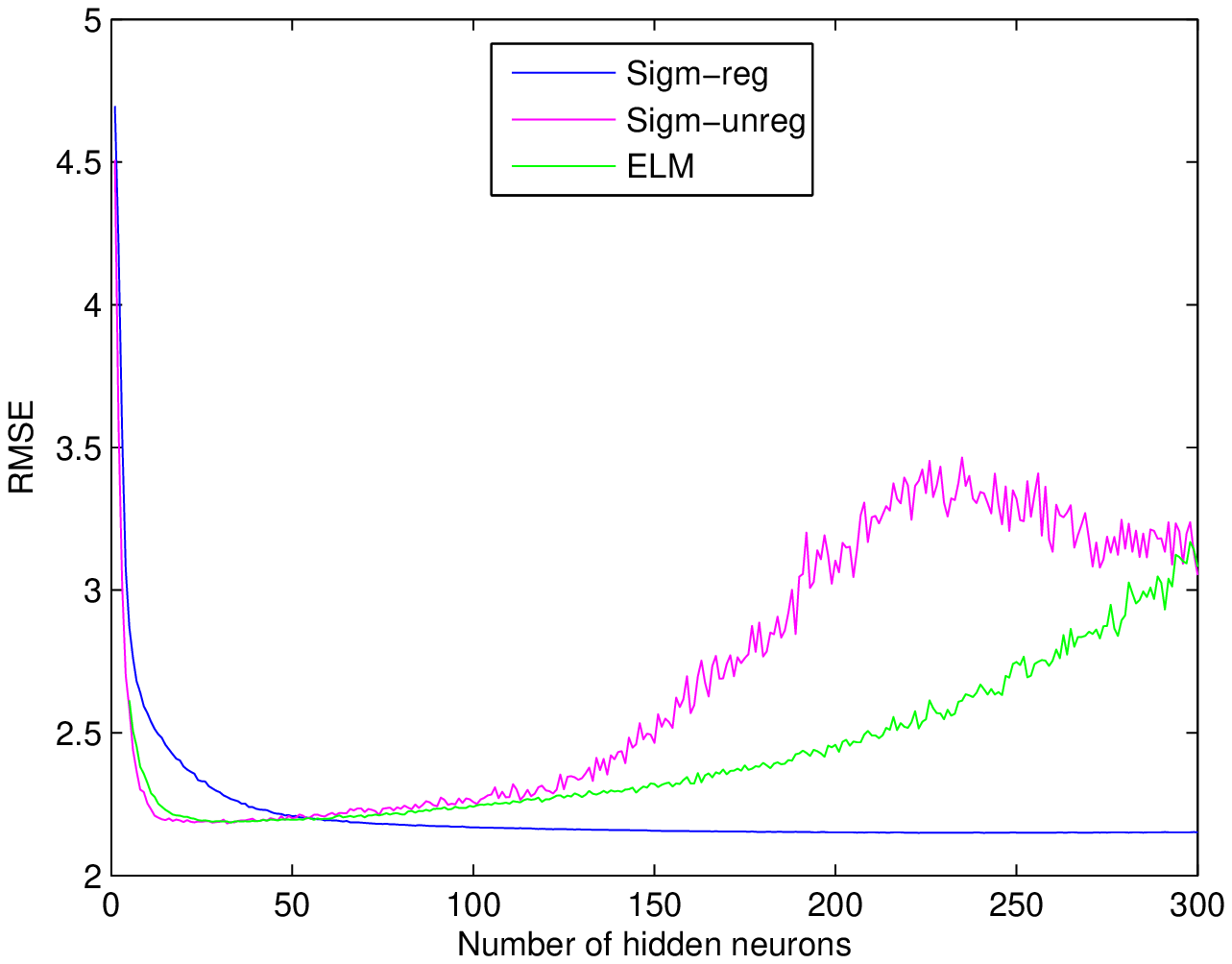}
\includegraphics[width=0.48\textwidth]{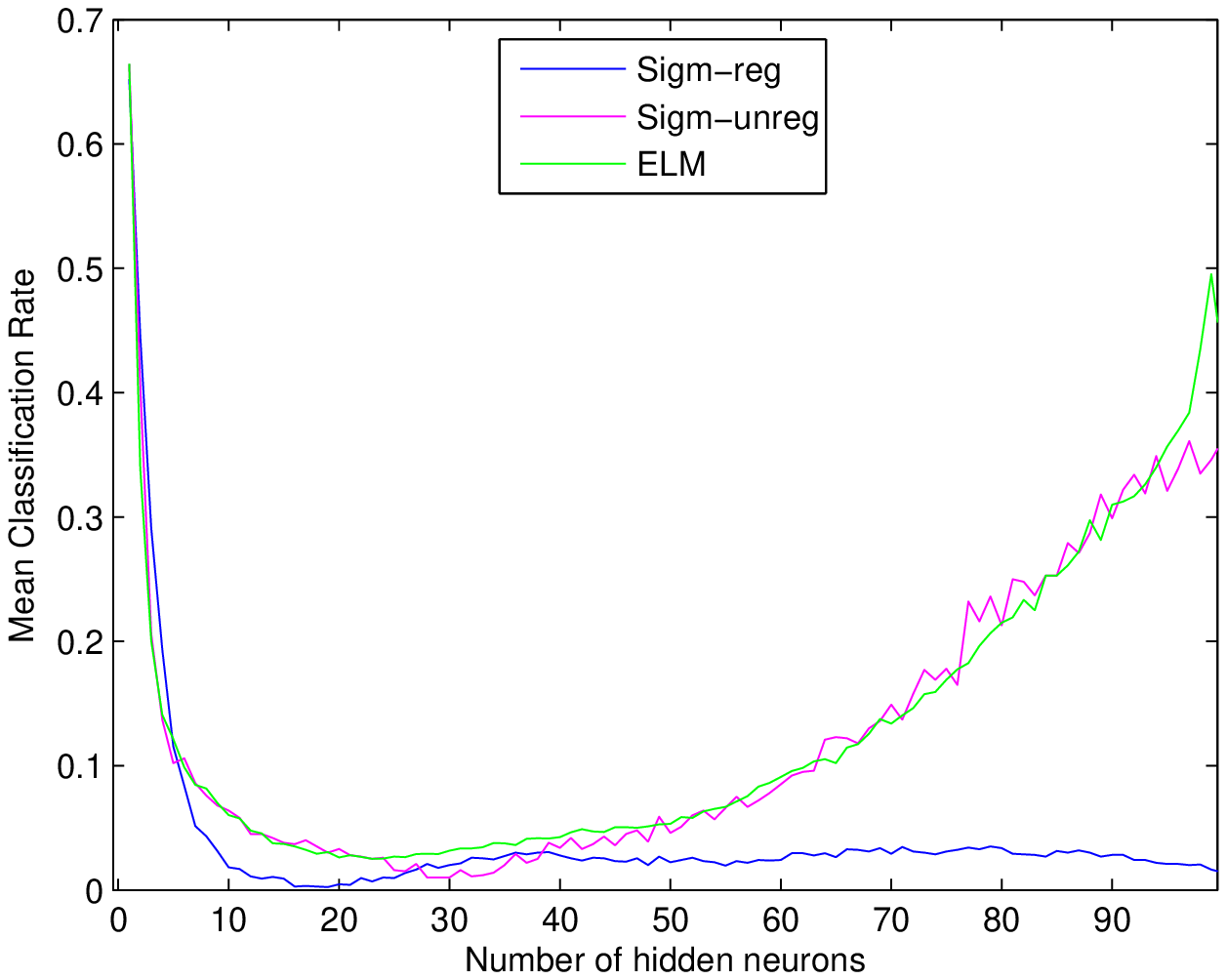}
\caption{Performance comparison in the Sigm (-reg and -unreg) cases: Abalone (Left); Iris (Right)}
\label{fig:PlotArray2}
\end{figure}

Figure \ref{fig:PlotArray} gives an insight on the 
performance trend (resp. average RMSE for regression or average 
misclassification rate for classification tasks) as a function of hidden space 
dimensionality for the cases HypT-reg, HypT-unreg and ELM. In Figure \ref{fig:PlotArray2} performance trends are shown for the cases SigmT-reg, Sigm-unreg and, for the sake of comparison, again ELM; because of their similarity with the HypT cases, plots are shown for only two datasets (i.e. Abalone and Iris).

It is interesting to note that when unregularised techniques are used, 
all datasets except Landsat show a fast error growth; 
besides, the curves have different characteristics for HypT-unreg and Sigm-unreg on one side and ELM on the other, showing error peaks in the formers while monotonically increasing error values are obtained in the latter. 
We conjecture the presence of two distinct phenomena: numerical instability and overfitting. 

In order to address the former issue, i.e. numerical instability, we evaluate for each dataset the ratio between the 
minimum singular value of hidden output matrix $ H $ and the Matlab 
default threshold (below which singular values are considered too small 
and therefore treated as zero). 
 
We checked that for each dataset processed with HypT-unreg or Sigm-unreg methods there is a {\it critical hidden layer size} above which the ratio becomes smaller than one; its trend is plotted in logarithmic units (red line, right scale), in Figure \ref{fig:PlotArray} for HypT-unreg case. 

When approaching critical size, inversion of singular values 
causes a wrong evaluation of $H^+$ and therefore a significant growth in the error; 
when the critical dimension is reached, singular values under threshold are automatically removed, thus allowing the subsequent decrease of error. The same trend was detected analysing the astronomical dataset in \citep{cancelliere2}.

This decrease is anyway not sufficient to reach optimal error values because 
of overfitting, which is known to arise when a large amount of hidden neurons 
is available to reproduce almost exactly the training data. 

An even more severe overfitting affects ELM in fact in this case 
test error is a monotonically increasing function of the number of hidden neurons. A possible explanation is that the setting of input weights in the interval (-1, 1) may allow `specialisation' of some hidden neurons on particular training instances, thus 
creating a sort of network partition, carried out thanks to saturation. 
On the contrary, when weights are randomly selected in the interval $ (-1/ \sqrt M, 1/ \sqrt M) $, as for HypT-unreg and Sigm-unreg cases, input weights are automatically kept small when the network size 
increases, thus exploiting the central part of both activation functions: consequently saturation is avoided \citep{cancelliere2}. 

After clarification of the main issues affecting unregularised approaches, 
we now discuss the regularised one, derived according to eq.(\ref{eq:tikmat}).

Looking at HypT-reg and Sigm-reg cases in Figures \ref{fig:PlotArray} and \ref{fig:PlotArray2}, 
it appears that not only numerical instability (i.e. the error peak) is removed, 
but also that the penalty term provides control of overfitting, avoiding error growth and allowing optimal exploitation of the superior potential of 
larger architectures. 
The error curves feature now a monotonic decrease, becoming increasingly 
smoother.

We obtained this result basing on the implications of eq.(\ref{eq:tiksing}) that clearly 
suggests the role of the $\lambda$ parameter in preventing instability: our original idea is to address the issue of its determination with 
an `ad hoc' tuning whenever 
growing error drift is experienced by the unregularised approach, as follows. 

We evaluate the validation error trend {\it inside the critical region}, looking for its minimum as a function of $ \lambda$; this allows to select a suitable value for this parameter for the subsequent experimentation 
with regularised pseudoinversion. 

We then made 100 random input weight choices and evaluated the 
mean test error (RMSE or misclassification rate) and the standard deviation $S$ for any number of hidden neurons.
In Table \ref{tabini} we list for HypT-reg and Sigm-reg the best ones of these mean values, called Optimal performance, together with the number of hidden neurons used to reach the associated performance (in parenthesis), and the corresponding standard deviation, as well as the selected value of $ \lambda$. 

Error values significantly better, basing on the Student's t-test evaluation of statistical confidence intervals, are recorded in bold. We can see that there is roughly no ``winner" between HypT-reg and Sigm-reg.

Some other interesting considerations can be made noting that the regularised error plots in Figures \ref{fig:PlotArray} and \ref{fig:PlotArray2}, after an initial rapid decrease, achieve a nearly constant regime, with 
small variation vs. increasing numbers of hidden neurons. 
We can thus define a ``near optimal" network size in both cases, as the one associated to 
a near optimal error level, i.e. an error significantly better than that from other methods. 
The assessment is again based on the Student's t-test evaluation of statistical confidence intervals. 

\begin{table}
\caption {Optimal performance for regression (top) and classification (bottom) datasets. For Delta Ailerons, average errors and standard deviations have to be multiplied by $10^{-4}$} 
\label{tabini}       
\footnotesize
\begin{tabular}{llllllllll} 
\hline\noalign{\smallskip}
                       & {\bf Abalone } &           &                  & {\bf Cpu}              &     &                  &  {\bf Delta Ail.}       &           & \\
                       &  RMSE           &  S        & $\lambda$   &  RMSE                  &  S  & $\lambda$   &  RMSE                  &  S        & $\lambda$ \\
HypT-reg          &   2.168 (192)    & 0.003  & $10^{-11}$  &  57.0 (78)   & 2.1 & $10^{-11}$ &   1.639 (147)         &  1.9e-3 & $10^{-13}$\\ 
Sigm-reg           & {\bf 2.150} (240)  & 0.004 & $10^{-12}$  &  57.3 (88)     & 1.8 & $10^{-10}$  &   {\bf{1.636}} (244) &  2.0e-3 & $10^{-13}$ \\ 
\noalign{\smallskip}
\hline\noalign{\smallskip}
& {\bf Iris }  &           &                & {\bf Diabetes}  &            &               &  {\bf Landsat} &            & \\
&  Err.(\%)   &  S(\%) & $\lambda$ &  Err.(\%)         &  S(\%) & $\lambda$ & Err.(\%)         &  S(\%) & $\lambda$ \\
HypT-reg  & $ \bf{0.02 }$ (66)   & 0.2 & $10^{-14}$ & $ 20.2 $ (347)  & 0.3 & $10^{-11}$ & \bf{12.65} (486) & 0.2  & $10^{-10}$ \\ 
Sigm-reg  &  0.2 (19)  & 1.1 & $10^{-14}$  & {\bf 19.4} (95) & 0.6 & $10^{-12}$  &  $ 13.77$ (396) & 0.3 & $10^{-11}$    \\
\noalign{\smallskip}
\hline\noalign{\smallskip}
\end{tabular}	
\end{table}

\begin{table}
\caption { Method comparison for regression (top) and classification (bottom) tasks. For Delta Ailerons, average errors and standard deviations have to be multiplied by $10^{-4}$} 
\label{tabmet1}       
\begin{tabular}{llclclc} 
\hline\noalign{\smallskip}
& {\bf Abalone } &  & {\bf Cpu}  &   &  {\bf Delta Ail.} &  \\
  &  RMSE  &  S  &  RMSE  &  S  &  RMSE  &  S   \\
HypT-reg (near opt.)&   {\bf{2.181}} (32)  & 0.011 & 57.55 (22)   & 6.2 &   {\bf{1.642}} (62) &  3.7e-3 \\ 
Sigm-reg (near opt.) & {\bf{2.181}} (74) & 0.008 & {\bf 57.05} (45) & 3.2 & 1.647 (99) & 3.2e-3 \\
\hline\noalign{\smallskip}
HypT-unreg	& 2.187 (34)  &  0.012 & 59.37 (17) &  7.2  &   1.649 (56)  & 2.8e-4   \\ 
Sigm-unreg & 2.183 (32)  &  0.013 & 57.58 (16) & 8.9 & 1.648 (45) & 6.5e-3 \\
ELM  & 2.186 (33)  & 0.015 & 59.48 (19) & 9.1 & 1.649 (75) &  8.8e-3 \\
\noalign{\smallskip}
\hline\noalign{\smallskip}
\hline\noalign{\smallskip}
& {\bf Iris }  &   & {\bf Diabetes}  &   &  {\bf Landsat} &  \\
  &  Err.(\%)  &  S(\%) &  Err.(\%) &  S(\%) & Err.(\%) &  S(\%)  \\
HypT-reg (near opt.)&  0.56 (37)   & 0.9  & 21.0 (146)  & 0.6 & 12.65 (486) & 0.2   \\ 
Sigm-reg (near opt.) & {\bf 0.3 } (16)  & 1.1 & {\bf20.7} (45) & 0.8  & 13.77 (396) & 0.3 \\
\hline\noalign{\smallskip}
HypT-unreg	& 1.02 (30)  &  1.1 & 21.2 (80) & 1.2 & 12.94 (295) & 0.4    \\
Sigm-unreg & 1.00 (29)  & 1.2  & 20.8 (86) & 1.4 & 12.78 (377) & 0.4 \\ 
ELM  &  2.52 (23)  & 1.9 & 21.2 (69) & 1.2 &  $ \bf{10.76 }$ (390) & 0.4    \\
\noalign{\smallskip}
\hline\noalign{\smallskip}
\end{tabular}	
\end{table}

Table \ref{tabmet1} compares the performance of this near optimal network with those obtained with the other methods, 
listing the best mean error (with the associated number of hidden neurons 
in parenthesis) and the corresponding standard deviation. 

Thus, it appears that regularisation provides, except for Landsat dataset, the best performance not only in terms of lowest error values (see table \ref{tabini})  but even with usage of smaller networks, limiting in this way computational 
cost and model complexity, and therefore fulfilling the goals set 
by previous researches (e.g. \citep{yu}). 
The smaller standard deviations almost always associated with the regularised
methods also suggest a lower dependence from initial conditions.

We also highlight that the use of small input weights and sigmoidal activation functions, which characterizes the Sigm-unreg case, allows to obtain error values lower with respect to the ELM case, so confirming the effectiveness of this choice in order to contain saturation and overfitting issues.

Landsat dataset constitutes an exception because best performance is 
reached using ELM. 
In this case, regularisation does not seem to be required, because 
overfitting and/or numerical instability do not take place, as it appears 
from unregularised error curves. 
The behaviour of the threshold ratio, which remains always larger than unity, 
is consistent with the lack of numerical instability and 
with our hypothesis of its relationship with error peaks. 
The lack of overfitting appears to be specific to the complexity of the dataset, 
having input vectors with size much larger than others, and therefore 
requiring a significantly larger number of parameters. 

In table \ref{tabtimes} are listed the computational times (in seconds) recorded for all datasets for completing one training step, i.e. one random setting of input weights and one output weights evaluation through pseudoinversion of the hidden layer output matrix; the number of hidden neurons has been fixed to $ 100 $ for the sake of comparison. 

We can see that, for each dataset, the times associated to each method do not differ significantly, because after having fixed the number of hidden neurons, the computational load necessary for the processing is  comparable.

The interested reader can find, for the common datatsets, a comparison among training times of ELM and backpropagation in \citep{huang}, and can verify that ELM turns out to be two or three orders of magnitude faster.

\begin{table}
\centering
\caption {Comparison of training times (s.) for regression (top) and classification (bottom) tasks at fixed hidden layer size (100 neurons).} 
\label{tabtimes}       
\begin{tabular}{llll} 
\hline\noalign{\smallskip}
& {\bf Abalone }  & {\bf Cpu}   &  {\bf Delta Ail.} \\
  &   &   &     \\
HypT-reg      &   0.032   & 0.035  &  0.093 \\ 
Sigm-reg      & 0.029 & 0.040 &  0.096  \\
HypT-unreg	  & 0.018   &  0.018  &   0.064  \\ 
Sigm-unreg  & 0.0151   &  0.025  &  0.077  \\
ELM            & 0.0123  & 0.015 &  0.060 \\
\noalign{\smallskip}
\hline\noalign{\smallskip}
& {\bf Iris }   & {\bf Diabetes}  &  {\bf Landsat}  \\
  &   &   &   \\
HypT-reg    &  0.037   &  0.0158  &  16.65   \\ 
Sigm-reg     & 0.032  &  0.0155 &  16.77\\
HypT-unreg	& 0.017 &  0.0139 &  15.94    \\
Sigm-unreg & 0.021  & 0.0144 &  15.78 \\ 
ELM          & 0.011  &  0.0116 &   14.32  \\
\noalign{\smallskip}
\hline\noalign{\smallskip}
\end{tabular}	
\end{table}

\section{Conclusions}

We have considered the numerical  instability ad overfitting problems for single hidden layer neural networks trained by pseudoinversion. 
We have shown how to use singular value analysis for the diagnosis of numerical instability, and how to solve this problem through  the determination of a critical hidden layer region from which the regularisation technique benefits. This method also contributes to reduce the overfitting. 

Tests have been performed for both regression and classification
tasks. For five out of six cases, the proposed regularisation  is proven necessary and  provides a significant performance improvement with respect to unregularised techniques; it also  allows to built lean architectures which achieve near optimal performance with a reduced number of hidden neurons. 

Moreover the use of sigmoidal activation functions and ``small" input weights (small because their values are linked to the hidden layer size), which characterizes the Sigm-unreg case, allows to obtain error values lower with respect to the ELM case, so confirming the effectiveness of this choice in order to contain saturation and overfitting issues.

Comparing our results on the common regression datasets 
with the alternative method proposed by Miche et al. \citep{Miche}, 
we note that our technique achieves RMSE values lower than those 
corresponding to their MSE values, with a somewhat lower number 
of neurons. 
Besides, in our opinion, our method is simpler, in the sense that 
it uses a single step of regularisation rather than two in their method, and we also deal with classification tasks.

\section{Acknowledgment}
The activity has been partially carried on in the context of the 
Visiting Professor Program of the Gruppo Nazionale per il Calcolo 
Scientifico (GNCS) of the Italian Istituto Nazionale di Alta 
Matematica (INdAM). 

\bibliographystyle{natbib}
\bibliography{mybibliography}

\end{document}